%% file: root.tex
\documentclass[letterpaper, 10 pt, conference]{ieeeconf} 

\IEEEoverridecommandlockouts

\usepackage{cite}
\usepackage{svg}
\usepackage{comment}
\usepackage{amsmath,amssymb,amsfonts}
\usepackage{algorithm}
\usepackage{algorithmicx}
\usepackage{algpseudocode}
\usepackage{graphicx}
\usepackage{textcomp}
\usepackage{float}
\usepackage{booktabs}
\usepackage{subcaption}

\newtheorem{assumption}{Assumption}

\usepackage{multirow}
\begin{document}
\title{\textbf{Switch-Time-Aware Hybrid Model Predictive Control\\ Using the Hybrid Minimum Principle}}
\author{\authorblockN{Valentin Gaucher and Wenlong Zhang}
\thanks{This work is supported by National Science Foundation under Grant No. 2331781.}
\thanks{The authors are with School of Manufacturing Systems and Networks, Ira A. Fulton Schools of Engineering, Arizona State University, Mesa, AZ, 85212, USA. Email: \{{\tt vgaucher, wenlong.zhang\}@asu.edu}.}}

\maketitle
\thispagestyle{empty} 
\pagestyle{empty}     
\overrideIEEEmargins 

\begin{abstract}
This article presents a time-aware hybrid model predictive control (MPC) framework that simultaneously optimizes the objective with respect to both continuous control inputs and event switching times. Rather than relying on a fixed time grid, the switching instant is lifted into the decision vector and treated as an explicit optimization variable. The resulting transcription is a smooth nonlinear program whose Karush–Kuhn–Tucker (KKT) conditions are consistent with the Hybrid Minimum Principle (HMP) event-time stationarity condition, so first-order hybrid sensitivities are recovered implicitly through the KKT system rather than by evaluating saltation matrices. Each MPC step solves a single sparse Quadratic Program (QP), jointly optimizing control inputs and impact timing in real time. On a bouncing mass landing on an oscillating platform, the method attaches in half the time of fixed-grid saltation MPC, reduces cumulative control effort by 30\%, and removes the post-impact rebounds, at a 15\% increase in per-cycle solve time. It matches the cumulative effort of offline time-aware solvers to within 4\% while running at 2.4 ms per cycle against 281 ms for a complementarity-based formulation of the same problem, of which a 10.6$\times$ reduction is attributable to the transcription alone at a fixed solver.
\end{abstract}

\input{1-Introduction}
\input{2-Preleminaries}
\input{3-ProblemFormulation}

\input{4-SQP}

\input{5-NumericalResults}
\input{6-Conclusion}

\bibliographystyle{IEEEtran}
\bibliography{IEEEabrv,refs}

\end{document}

%% file: 1-Introduction.tex
\section{INTRODUCTION}
In this paper, we address Model Predictive Control (MPC) for hybrid systems subject to state-dependent, instantaneous resets within the prediction horizon. In these systems, the event time is a critical decision variable coupled with the continuous states and inputs \cite{ALTIN2018128}. A primary challenge in real-time execution is handling state-dependent resets. Standard fixed-step Jacobians neglect the sensitivity of the state trajectory to the exact switching time, yielding suboptimal mode-switching decisions \cite{KongProcIEEE2024, KongMPC}. The proposed approach in this paper converts the implicit dependence of switching time on the state trajectory into an explicit optimization variable, allowing standard nonlinear programming (NLP) solvers to handle hybrid sensitivity information without differentiating through discontinuities.

While rigorous frameworks address this complexity, they are often computationally demanding for real-time applications. Mixed-Integer QP (MIQP) \cite{MIQPSanfelice, BEMPORAD2018412} uses binary variables to optimize timings, Mathematical Programs with Complementarity Constraints (MPCC) \cite{Baumrucker2009} optimize across transitions, and move finite-element meshes \cite{KaziWangBiegler2025} precisely align switching surfaces. Despite robust embedded solvers for these formulations \cite{BEMPORAD2018412}, they rely on combinatorial searches or large-scale formulations that can be prohibitive in real time, while standard real-time MPC instead fixes the switching time or ignores first-order timing sensitivities. This gap calls for a formulation that achieves the timing precision of MPCC or variable-step methods while preserving the speed of fixed-step Sequential Quadratic Programming (SQP) solvers.

\vspace{0.02in}
\noindent \textbf{Related Work.}
The theoretical conditions for optimality in hybrid systems are well-established. Sussmann \cite{Sussmann1999} and Caines \cite{Caines2021} extended the Pontryagin Maximum Principle to hybrid systems, deriving conditions that optimize both continuous controls and discrete switching times. Dmitruk and Kaganovich \cite{DmitrukKaganovich2008} unified these under the Hybrid Minimum Principle (HMP). While mathematically powerful, the HMP is traditionally a continuous-time, open-loop theory. Enforcing these optimal conditions inside a discrete-time, receding-horizon feedback loop remains challenging.

Numerical transcriptions bridge the gap between continuous theory and discrete solvers. Ross and Fahroo \cite{RossFahroo2004} demonstrated that for smooth systems, KKT stationarity conditions and their multipliers in pseudospectral methods converge to Pontryagin costates. Indirect methods, such as the recent work by Hu et al. \cite{HuHMP2025}, solve HMP conditions directly via successive approximation, but are designed for open-loop trajectory generation rather than closed-loop control.

To incorporate switching effects into gradient-based controllers, one must differentiate through the discrete transition. This is achieved via the saltation matrix, which corrects the sensitivity analysis across a transversal guard. Formalized by Aizerman et al. \cite{Aizerman1958} and further developed in foundational works \cite{Filippov1988, leine2003}, Kong et al. \cite{KongProcIEEE2024} recently showed the practical utility of these matrices for robotics. However, standard discrete-time implementations typically treat the switching time as a passive consequence of the forward rollout, rather than an active decision variable.

\vspace{0.02in}
\noindent \textbf{Contributions.}
This paper presents HMP-MPC, a framework that incorporates Hybrid Minimum Principle (HMP) consistency within an SQP-based Real-Time Iteration (RTI) scheme. The idea is lifting the switching instant as an explicit decision variable, transforming the nonsmooth dependence of the dynamics on event timing into a smooth NLP with algebraic constraints. This enables standard gradient-based solvers to optimize impact timing directly. The contributions are: \textit{(i)} Lifting the switching time as an intra-step decision variable, localizing temporal adjustments to each event interval while preserving the $\mathcal{O}(N)$ sparse block-tridiagonal structure. \textit{(ii)} A smooth NLP transcription of hybrid transitions using algebraic equality constraints, yielding a $C^1$ representation suitable for SQP-based optimization. \textit{(iii)} Implicit recovery of first-order hybrid sensitivities through the KKT system, yielding conditions consistent with the HMP time-stationarity relation.

%% file: 2-Preleminaries.tex
\section{PRELIMINARIES}

\subsection{Notation}
Let $\mathbb{R}$ denote the set of real numbers, $\mathbb{R}_{\geq 0}$ the set of non-negative real numbers, and $\mathbb{N}$ the set of natural numbers. Given a continuously differentiable function $f$, $\nabla_x f$ denotes its gradient with respect to $x$. The space of $k$-times continuously differentiable functions is denoted $C^k$.

Trajectories are parameterized by a hybrid time domain $E \subset \mathbb{R}_{\geq 0} \times \mathbb{N}$, where $t \in \mathbb{R}_{\geq 0}$ denotes the continuous flow time and $j \in \{1, 2, ..., J\}$ denotes the discrete jump counter. A hybrid time domain $E$ is defined as a finite union of intervals:
\begin{equation}
    E = \bigcup_{j=0}^{J-1} [t_j, t_{j+1}] \times \{j\},
\end{equation}
where $0 = t_0 \leq t_1 \leq \dots \leq t_J$. For a generic time-dependent state trajectory $x(t,j)$, we define the left limit (pre-jump state) at the transition instant $t_j$ as $x^-=x^-(t_j, j) := \lim_{t \nearrow t_j} x(t, j)$, and the right limit (post-jump state) as $x^+=x^+(t_j, j+1) := x(t_j, j+1)$. The same definitions apply to the control input, denoted $u$, $u^-$, and $u^+$.

\subsection{Hybrid Optimal Control Problem}
We consider a nonlinear, controlled hybrid system $\mathcal{H}$ defined by the tuple $(X, U, \mathcal{Q}, F, G, R)$. The continuous state space is $X \subseteq \mathbb{R}^n$, the control input space is $U \subseteq \mathbb{R}^m$, and $\mathcal{Q}$ is a finite set of discrete modes. The system dynamics are partitioned into continuous flows and discrete jumps:
\begin{align}\label{eq:hybridSys}
\begin{split}
    \dot{x}(t,j) = f_{q}(x(t,j), u(t,j)),\qquad & (x,u) \in F \\
    x^+(t_j,j+1) = R(x^-(t_j,j), t_j),\qquad & x \in G
\end{split}
\end{align}

The continuous evolution is governed by the $C^1$ vector field $f := \{f_q\}_{q \in \mathcal{Q}}$ operating within the flow set $F \subseteq X \times U$. Discrete transitions occur when the trajectory intersects the jump set $G \subseteq X$, defined as the zero-level set $G := \{ x \in X \mid g(x,t) = 0 \}$ of a $C^2$ guard function $g: X \times \mathbb{R}_{\geq 0} \to \mathbb{R}$. When $x^-(t_j, j) \in G$, a transition $\sigma: q \to r$ is triggered and the state is instantaneously updated via the $C^2$ reset map $R: X \times \mathbb{R}_{\geq 0} \to X$. 

\begin{assumption}[Transversality and Regularity]\label{assump:regularity}
    The state trajectory strictly crosses the jump set $G$ and does not graze it. Specifically, the transversality condition holds for all interior events at the switching instant $t_j$:
    \begin{equation}\label{eq:transversality}
        \nabla_x g(x^-,t_j)^\top f_{q}(x^-,u^-)+\nabla_t g(x^-,t_j) \neq 0.
    \end{equation}
    Furthermore, the control set $U$ is convex and compact, and the Linear Independence Constraint Qualification \cite{Henrion1992} holds for all optimization constraints.
\end{assumption}
\vspace{0.04in}

We formulate a continuous-time optimal control problem for $\mathcal{H}$ over a fixed discrete horizon $J \in \mathbb{N}_{>0}$, which fixes the number of physical jumps in the prediction window, with $t_J$ the free terminal time. The objective is to minimize:
\begin{align}\label{eq:Bolza}
\begin{split}
    \mathcal{J} &= \sum_{j=0}^{J-1} \int_{t_j}^{t_{j+1}} L_C(x(t,j), u(t,j)) dt\\
    &\quad+ \sum_{j=1}^{J} L_D(x^-(t_j,j), t_j) + V(x(t_J, J)),
\end{split}
\end{align}
where $L_C\geq0$ represents the $C^1$ continuous flow cost, $L_D\geq0$ represents the discrete transition cost incurred at the jump, and $V$ is the terminal cost.
Since $L_D$ is only evaluated at $(x^-(t_j,j), t_j)$, we will use $L_D:=L_D(x^-(t_j,j), t_j)$ for the rest of the derivations. Similarly we use $R:= R(x^-(t_j,j), t_j)$ and $g:=g(x^-(t_j,j), t_j)$.

\subsection{The Hybrid Minimum Principle (HMP)}
The HMP provides the necessary conditions for optimality for the defined hybrid system \cite{Caines2021}. Let $\lambda \in \mathbb{R}^n$ denote the continuous costate (adjoint) vector. The Hamiltonian in mode $q$ is defined as $H_q(x,u,\lambda) := L_q(x,u) + \lambda^\top f_q(x,u)$. 

At a switching instant $t_j$, intersection with the jump set $G$ induces a discrete discontinuity governed by a Lagrange multiplier $\kappa_j$ associated with the active guard constraint $g(x,t)=0$:
\begin{subequations}
\begin{align}
    \lambda^- &= \nabla_x R^{\top} \lambda^+  + \kappa_j\nabla_x g +\nabla_x L_D \label{eq:costate_jump}\\
    H_q^- &= H_r^+ - \nabla_t L_D - \kappa_j \nabla_t g - (\lambda^+)^\top \nabla_t R \label{eq:hamiltonHMP}
\end{align}
\end{subequations}
where $H_{q}^- := H_{q}(x^-, u^-, \lambda^-)$ and $H_{r}^+ := H_{r}(x^+, u^+, \lambda^+)$. Equation \eqref{eq:hamiltonHMP} encodes the event-time stationarity condition.

\subsection{First-Order Sensitivities and the Saltation Matrix}
The first-order sensitivity of the post-jump state $\delta x^+$ with respect to the pre-jump state $\delta x^-$ is mathematically captured by the saltation matrix $\Xi$ \cite{KongProcIEEE2024}, providing the first order mapping $\delta x^+=\Xi \delta x^-$. It captures how a perturbation in the pre-impact state shifts both the impact time and the post-impact trajectory, and is expressed as:
\begin{equation} \label{eq:saltation}
    \Xi = \nabla_x R + \frac{(f_r^+ - \nabla_x R f_q^- - \nabla_t R)\, \nabla_x g^\top}{\nabla_x g^\top f_q^- + \nabla_t g}
\end{equation}
where $f_q^- := f_q(x^-, u^-)$ and $f_r^+ := f_r(x^+, u^+)$.
All derivatives of $R$ and $g$ are evaluated at $(x^-, t_j)$ unless otherwise specified. The denominator represents the total time derivative of the guard ($D_t g$), which is non-zero by the transversality condition \eqref{eq:transversality}.

%% file: 3-ProblemFormulation.tex
\section{SWITCH-TIME-AWARE TRANSCRIPTION}

Standard fixed-grid MPC treats impact timing as an uncontrollable consequence of the state trajectory. By lifting the switching time into the decision vector, our formulation allows the controller to optimize the trade-off between continuous control effort and the event timing (Fig. \ref{fig:LiftedTime}). We introduce the event-localized discretization, formulate the discrete optimal control problem, and prove that its KKT conditions recover the first-order sensitivities of the HMP.

\subsection{Event-Localized Discretization}
To solve the optimal control problem numerically, we discretize the hybrid time domain. While general adaptive time-mesh refinement algorithms have been proposed to improve discretization accuracy in continuous-time MPC \cite{PAIVA2018104}, we use an event-localized scheme. 
The derivation uses a first-order Forward Euler scheme to expose the analytical equivalence between the KKT conditions and the continuous time HMP.

\begin{figure}[h]
    \centering
    \includegraphics[width=0.92\linewidth]{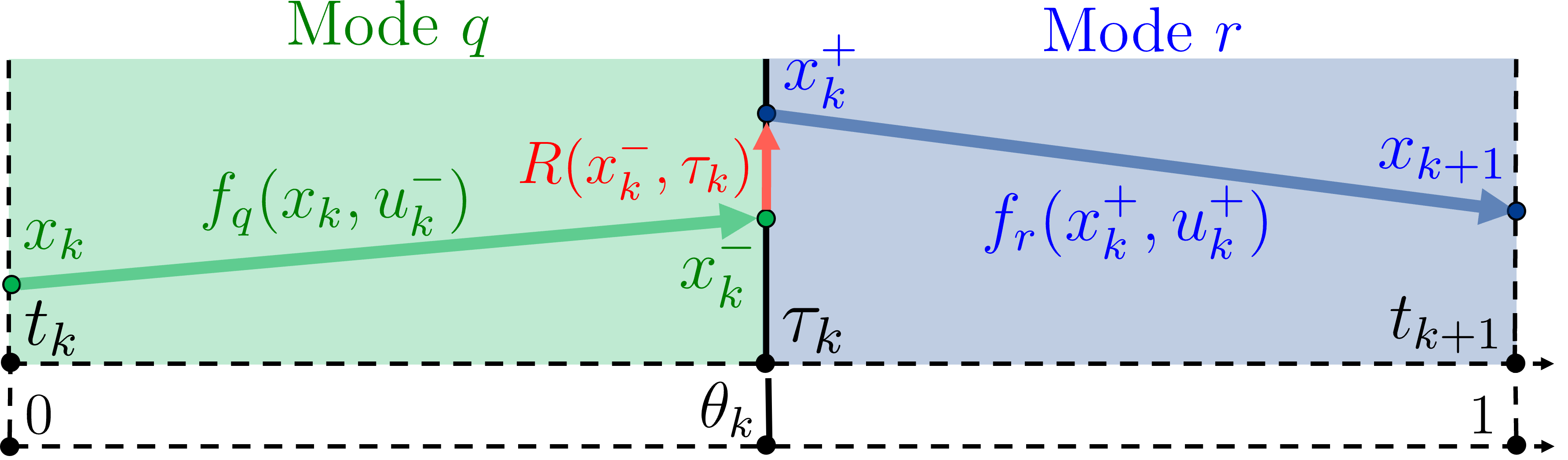}
    \caption{State-space representation of the event-localized discretization.}
    \label{fig:reset}
\end{figure}

We partition the horizon into a discrete grid indexed by $k \in \{0, \dots, N-1\}$ with uniform steps $h_k = t_{k+1}-t_k$. We restrict the formulation to at most one discrete transition per interval. The switching time is parameterized by a continuous intra-step fraction $\theta_k \in (0,1)$, defining the switching instant as $\tau_k := t_k + \theta_k h_k$ (see Fig.~\ref{fig:reset}). Therefore, $\theta_k$ allows the optimizer to `slide' the impact within the time step, rather than accepting the discretization error imposed by a fixed grid, and decouples event timing from state propagation so that timing is adjusted without implicitly differentiating through the guard (the guard constraint enforces consistency). The strict interior constraint enforces the transversality condition of Assumption 1 numerically.

\begin{figure*}[h]
    \centering
    \begin{subfigure}{0.51\linewidth}
        \centering
        \includegraphics[width=\linewidth]{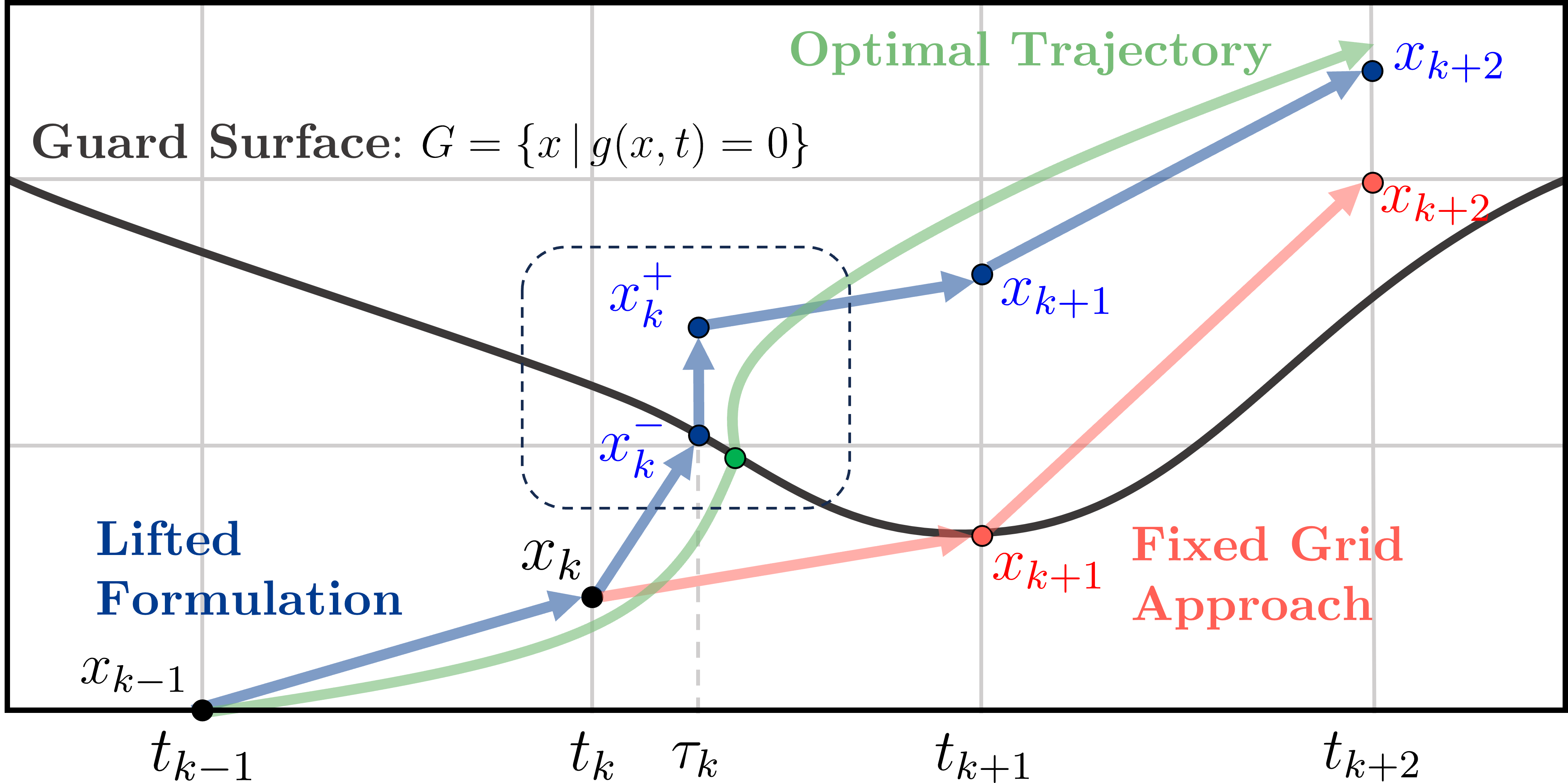}
        \caption{Lifted formulation (blue) localizes the reset within $[t_k, t_{k+1}]$ at $\tau_k$, while the fixed-grid approach (red) defers it to the grid node $t_{k+1}$,  accumulating a timing error over subsequent steps.}
        \label{fig:LiftedTime}
    \end{subfigure}
    \hfill
    \begin{subfigure}{0.41\linewidth}
        \centering
        \includegraphics[width=\linewidth]{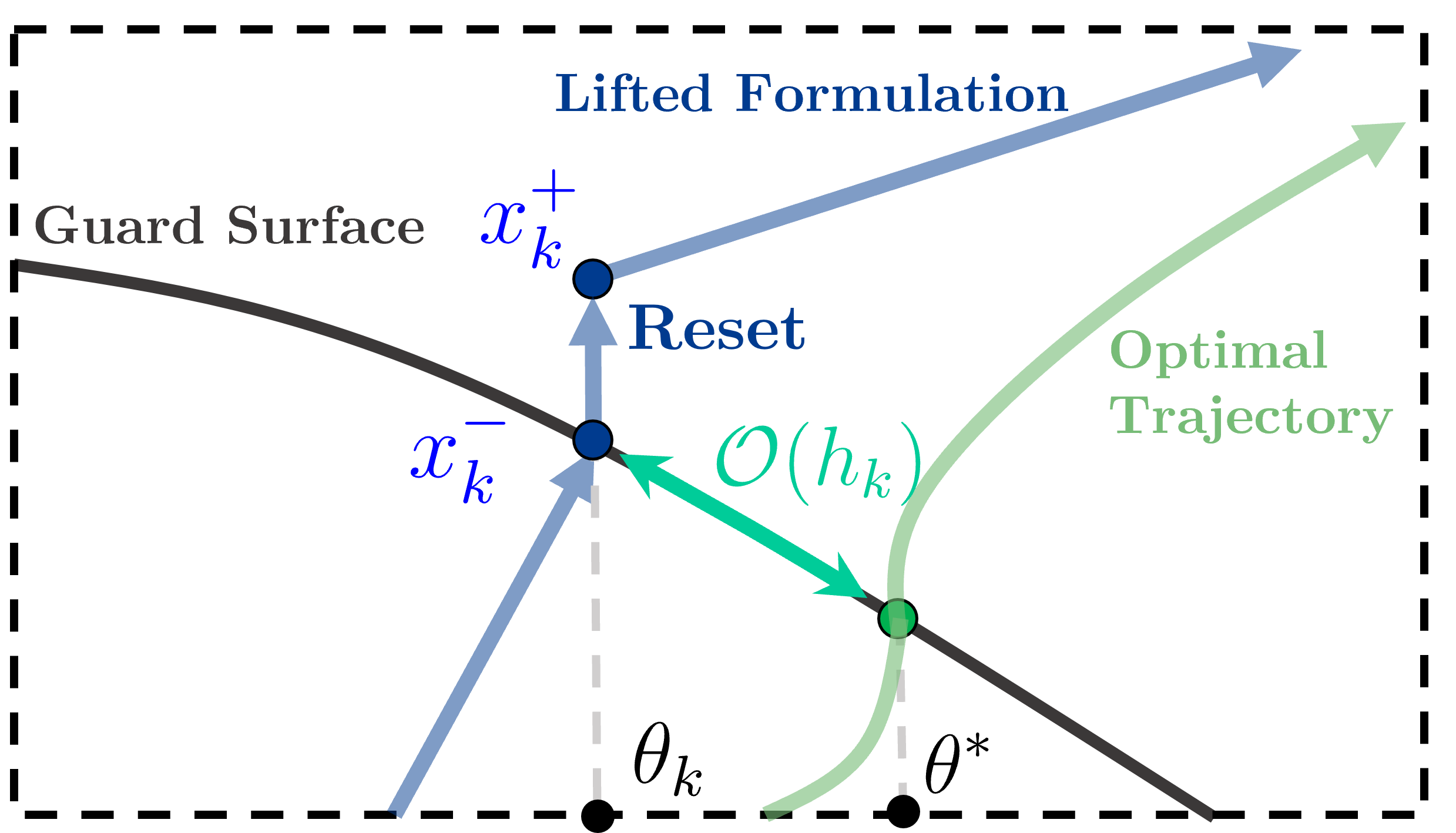}
        \caption{Zoom at the guard intersection. The lifted formulation places $x_k^-$ on the guard at $\tau_k = t_k + \theta_k h_k$, with the discrete optimum $\theta_k$ lying $\mathcal{O}(h_k)$ from the continuous-time HMP optimum $\theta^*$.}
        \label{fig:LiftedTime2}
    \end{subfigure}
    \caption{Event-localized transcription: the lifted formulation applies the reset at the intra-step instant $\tau_k$, whereas the fixed-grid approach defers it to the node $t_{k+1}$.}
    \label{fig:LiftedTimeAll}
\end{figure*}

The pre-event arc of length $h_k\theta_k$ evolves the state strictly to the guard surface:
\begin{subequations}
\begin{equation}\label{eq:Fminus}
    x_k^- = x_k + h_k\theta_k f_{q_k}(x_k,u_k^-).
\end{equation}
If the guard condition is met ($g(x_k^-,\tau_k)=0$), the state undergoes the discrete transition $x_k^+ = R(x_k^-,\tau_k)$ to mode $r_k$. The post-event arc of length $h_k(1-\theta_k)$ then propagates the state to the subsequent discrete node:
\begin{equation}\label{eq:Fplus}
    x_{k+1} = x_k^+ + h_k(1-\theta_k) f_{r_k}(x_k^+,u_k^+).
\end{equation}
\end{subequations}
The proposed transcription constitutes a first-order consistent discretization of the hybrid optimal control problem \eqref{eq:Bolza} under the forward Euler scheme (Fig. \ref{fig:reset} \& \ref{fig:LiftedTime}).

\subsection{Optimal Control Problem Formulation}
Let $\mathcal{K}\subseteq \{0, \dots, N-1 \}$ denote the set of discrete grid intervals during which a physical jump occurs. 
\begin{equation}\label{eq:K}
    \mathcal{K} := \{ k \mid (x_k^-, \tau_k) \in G, \; \tau_k \in ]t_k, t_{k+1}[\, \}.
\end{equation}

Using the decision vector $z := \left[x_0,\ \{u_k^-, u_k^+, \theta_k\}_{k \in \mathcal{K}},\ \{u_k\}_{k \notin \mathcal{K}},\ x_N \right]$  where each transition interval $k\in\mathcal{K}$ contributes an intra-step fraction $\theta_k$ and a split input pair $(u_k^-, u_k^+)$ while non-transition intervals retain a single input $u_k$. We minimize the objective function $\mathcal{J}(z)$ (the sum of the continuous flow costs and discrete jump costs) over the prediction horizon:
\begin{align} \label{eq:discret-cost}
\begin{split}
    \mathcal{J}(z) &=\sum_{k\notin \mathcal{K}} h_k L_{C,q_k}(x_k,u_k)+\\
    &\quad\sum_{k\in\mathcal{K}} \Big( h_k L_{C,\sigma_k}(z_k) + L_D(x_k^-, \tau_k) \Big) + V(x_N)
\end{split}
\end{align}
where $q_k\in \mathcal{Q}$ denotes the active discrete mode at interval $k$, and 
\begin{equation}
    L_{C,\sigma_k}= \theta_k L_{C,q_k}(x_k,u_k^-) + (1-\theta_k) L_{C,r_k}(x_k^+,u_k^+)
\end{equation}
is the two-arc continuous flow cost at transition interval $k$. Each product $h_kL_{C,(\cdot)}$ is the left-endpoint rectangle approximation of $\int_{t_k}^{t_{k+1}}L_Cdt$ in \eqref{eq:Bolza}, consistent with the Forward Euler. The minimization is subject to the discrete forward-Euler dynamic constraints \eqref{eq:Fminus} and \eqref{eq:Fplus}, the algebraic guard condition $g(x_k^-,\tau_k) = 0$, the reset map $x_k^+ = R(x_k^-, \tau_k)$, and the compact state-input bounds $x \in X$, $u \in U$.

For ease of notation during this event-localized analysis, we drop the interval subscript on the discrete modes and denote the pre-event mode $q_k$ as $q$, and the post-event mode $r_k$ as $r$. We construct the discrete Lagrangian for a generic interval $k\in\mathcal{K}$ containing a transition. Assigning the dual variables $\lambda_k^-, \lambda_k^+, \lambda_{k+1} \in \mathbb{R}^n$ and $\kappa_k \in \mathbb{R}$, the Lagrangian $\mathcal{L}_k$ is defined as:
\begin{align*}\label{eq:Lagrangian}
    \mathcal{L}_k = \, & h_k L_{C,\sigma_k} + L_D(x_k^-, \tau_k) 
    - (\lambda_k^-)^\top \big(x_k^- - x_k - h_k\theta_k f_{q}^-\big) \\
    &+ \kappa_k\, g(x_k^-, \tau_k) 
    - (\lambda_k^+)^\top \big(x_k^+ - R(x_k^-, \tau_k)\big)  \\
    &- (\lambda_{k+1})^\top \big(x_{k+1}-x_k^+ - h_k(1-\theta_k) f_{r}^+\big)\addtocounter{equation}{1}\tag{\theequation}
\end{align*}

The pre and post event inputs $u_k^-$ and $u_k^+$ are treated as independent decision variables, each subject to the same compact bound $U$, with no continuity or hybrid constraints imposed across the jump.

\subsection{Recovery of Hybrid Sensitivities}
In this subsection, we will use the following notations. The event-local Hamiltonians $H_{q}^- = H_{q}(x_k^-,\lambda_k^-,u_k^-)$ and $H_{r}^+=H_{r}(x_k^+,\lambda_k^+,u_k^+)$, as well as the post-arc flow Hamiltonian $H_{r}=H_{r}(x_k^+,\lambda_{k+1},u_k^+)$. All derivatives of $R$, $g$, and $L_D$ are evaluated at $(x_k^-, \tau_k)$ unless otherwise specified.

\vspace{0.02in}
\textbf{Theorem 1.} Consider the optimal control problem defined by \eqref{eq:discret-cost} with dynamics \eqref{eq:Fminus}--\eqref{eq:Fplus} under Assumption 1. Let $f_q, g, R, L_{C,q},$ and $L_D$ be continuously differentiable. Any Karush-Kuhn-Tucker (KKT) point of this problem satisfies a discrete adjoint jump and event-time stationarity condition:
\begin{subequations}
\begin{equation}\label{eq:thm_adjoint}
    \lambda_k^- = \nabla_x R^\top \lambda_k^+ + \kappa_k \nabla_x g + \nabla_x L_D,
\end{equation}
\begin{equation}\label{eq:thm_stationarity}
    H_q^- = H_r^+ -  \nabla_t L_D - \kappa_k  \nabla_t g - (\lambda_k^+)^\top \nabla_t R + \mathcal{O}(h_k)
\end{equation}
\end{subequations}
constituting a first-order consistent discretization of the generalized HMP conditions. 

\vspace{0.02in}
\textbf{Proof.}
We analyze the KKT conditions of the discrete Lagrangian $\mathcal{L}_k$ \eqref{eq:Lagrangian}. For the \textit{costate jump condition,} setting $\nabla_{x_k^-} \mathcal{L}_k = 0$ yields
\begin{equation}\label{eq:A-}
    \lambda_k^- = \nabla_x R^\top \lambda_k^+ + \kappa_k \nabla_x g + \nabla_x L_D,
\end{equation}
which coincides with the HMP adjoint jump and recovers \eqref{eq:thm_adjoint}. Similarly, computing the condition $\nabla_{x_k^+}\mathcal{L}_k=0$ gives
\begin{equation}\label{eq:A+}
     \lambda_k^+ = \lambda_{k+1} + h_k(1-\theta_k)\,\nabla_{x_k^+} H_{r}.
\end{equation}

\noindent\textit{Event-time stationarity.} Setting $\nabla_{\theta_k} \mathcal{L}_k = 0$ isolates the sensitivity to the event time. Since $x_k^-$ is independent from $\theta_k$ due to the lifted parametrization, $x_k^-$ is treated as an independent decision variable constrained by \eqref{eq:Fminus}, giving $\nabla_{\theta_k}x_k^- = 0$. The jump cost $L_D$ explicitly depends on $\tau_k = t_k + \theta_k h_k$. Applying the chain rule yields $\nabla_{\theta_k} L_D(x_k^-, \tau_k) = h_k \nabla_t L_D$. Dividing the stationarity condition by $h_k$ yields:
\begin{equation} \label{eq:proof_stationarity_raw}
    H_q^- = H_r - \nabla_t L_D - \tfrac{\kappa_k}{h_k} \nabla_{\theta_k} g - \tfrac{1}{h_k}(\lambda_k^+)^\top \nabla_{\theta_k} R.
\end{equation}
From \eqref{eq:A+} and the Hamiltonian definition, $H_r = H_r^+ - h_k(1-\theta_k)\nabla_x H_r f_{r}^+ = H_r^+ +\mathcal{O}(h_k)$. Therefore, substituting $H_r^+$ and applying the chain rule to both $\nabla_{\theta_k} g$ and $\nabla_{\theta_k} R$ allows \eqref{eq:proof_stationarity_raw} to be reformulated as \eqref{eq:thm_stationarity} and matching the HMP condition to first order:
\begin{equation}\label{eq:thm_stationarity_proof}
    H_q^- = H_r^+ - \nabla_t L_D - \kappa_k \nabla_t g - (\lambda_k^+)^\top \nabla_t R + \mathcal{O}(h_k).
\end{equation}
This completes the proof. $\hfill \blacksquare$

The $\mathcal{O}(h_k)$ term depends on the local truncation error of the integration scheme and remains bounded under standard Lipschitz assumptions on $f,\ g$ and $R$. In the limit $h_k \to 0$ the KKT system formally recovers the continuous-time HMP conditions. Fig. \ref{fig:LiftedTime2} illustrates this geometrically and Fig.~\ref{fig:stationarity} numerically.

\vspace{0.02in}
\textbf{Corollary 1.} Under the conditions of Theorem 1, eliminating the event-time Lagrange multiplier $\kappa_k$ from the KKT system yields the costate update:
\begin{equation} \label{eq:thm_saltation}
    \lambda_k^- = \Xi_k^\top \lambda_k^+ + \nabla_x L_D - \frac{D_t L_D}{D_t g}\nabla_x g + \mathcal{O}(h_k),
\end{equation}
where $\Xi_k$ is the standard saltation matrix \eqref{eq:saltation}, $D_t L_D := \nabla_x L_D^\top f_q^- + \nabla_t L_D$ is the total time derivative of the cost along the pre-event vector field, and $D_t g := \nabla_x g^\top f_q^- + \nabla_t g$. 

\vspace{0.02in}
\textbf{Proof.}
\textit{(For clarity of exposition, we take $L_{C,q} = L_{C,r}$. The fully generalized result for discontinuous flow costs is given in Remark 1).}
Solving the stationarity condition \eqref{eq:thm_stationarity_proof} for $\kappa_k$ requires expanding the Hamiltonians $H_q^- = (\lambda_k^-)^\top f_q^- + L_{C,q}$ and $H_r^+ = (\lambda_k^+)^\top f_r^+ + L_{C,r}$. Substituting the adjoint jump \eqref{eq:A-} into $H_q^-$ and equating to \eqref{eq:thm_stationarity_proof} gives:
\begin{align} \label{eq:proof_kappa}
\begin{split}
    \kappa_k &= \frac{(\lambda_k^+)^\top \big(f_r^+ - \nabla_x R f_q^- - \nabla_t R \big)}{\nabla_x g^\top f_q^- + \nabla_t g}\\
    &\quad - \frac{\nabla_x L_D^\top f_q^- + \nabla_t L_D}{\nabla_x g^\top f_q^- + \nabla_t g}+ \mathcal{O}(h_k).
\end{split}
\end{align}
By Assumption 1, the transversality condition ensures $D_t g\neq 0$ at the interior event, guaranteeing $\kappa_k$ remains bounded. Substituting Eq. \eqref{eq:proof_kappa} into the adjoint jump condition \eqref{eq:A-} yields:
\begin{align}
\begin{split}
    \lambda_k^- &= \left( \nabla_x R + \tfrac{(f_r^+ - \nabla_x R f_q^- - \nabla_t R)\nabla_x g^\top}{D_t g} \right)^\top \lambda_k^+ \\
    & \quad+ \nabla_x L_D - \frac{D_t L_D}{D_t g}\nabla_x g + \mathcal{O}(h_k).
\end{split}
\end{align}
Recognizing the saltation matrix $\Xi_k$ multiplying $\lambda_k^+$, we obtain \eqref{eq:thm_saltation}, completing the proof. $\hfill \blacksquare$

This corollary shows that our formulation performs the saltation correction through the KKT multipliers, recovering the same sensitivity that explicit saltation methods compute directly.

\vspace{0.02in}
\noindent\textbf{Remark 1.} 
If the flow cost is discontinuous across the jump ($L_{C,q} \neq L_{C,r}$), the event-time stationarity condition captures this variation $\Delta L_{C} = L_{C,r} - L_{C,q}$. Following identical steps without canceling $L_{C,q}$ and $L_{C,r}$ yields the fully generalized update:
\begin{align}
\begin{split}
    \lambda_k^- &= \Xi_k^\top \lambda_k^+ + \nabla_x L_D - \tfrac{D_t L_D - \Delta L_{C}}{D_t g} \nabla_x g + \mathcal{O}(h_k).
\end{split}
\end{align}

\vspace{0.02in}
\noindent\textbf{Remark 2.} The $\mathcal{O}(h_k)$ remainder  in \eqref{eq:thm_stationarity} originates from 
the approximation $H_r = H_r^+ + \mathcal{O}(h_k)$ in  \eqref{eq:A+}, which reflects the local truncation error of 
the Forward Euler scheme. For any $p$-th order consistent  integrator, this approximation tightens to $H_r = H_r^+ + 
\mathcal{O}(h_k^p)$, and the remainder in Theorem 1 and Corollary 1 improves accordingly. Furthermore, the guard and reset constraints are evaluated at the boundary $(x_k^-, \tau_k)$ and are decoupled from the integration scheme, so the formulation is directly compatible with higher-order methods.

%% file: 4-SQP.tex
\section{REAL-TIME ITERATION FOR THE LIFTED HYBRID NLP}

\subsection{The Linear-Quadratic Subproblem}
We embed the lifted hybrid NLP within a real-time iteration (RTI) scheme \cite{Diehl2005}, solving one structured QP per control cycle. The nonlinear problem is linearized around a nominal trajectory from the previous solver iteration, denoted by $\tilde{z}$. We formulate a sparse QP to compute the optimal perturbation vector $\Delta z=z-\tilde{z}$. 

To define the affine dynamic constraints of the QP, let $A_q := \nabla_x f_q(\tilde{x}, \tilde{u})$ and $B_q := \nabla_u f_q(\tilde{x}, \tilde{u})$ denote the Jacobians of the continuous vector field in mode $q$. The pre-event arc dynamics \eqref{eq:Fminus} are linearized as:
\begin{equation} \label{eq:lin_dyn_minus}
    \Delta x_k^- = \Phi_q \Delta x_k + \Gamma_q \Delta u_k^- + S_q \Delta \theta_k,
\end{equation}
where $\Phi_q = (I + h_k \tilde{\theta}_k A_q) \in \mathbb{R}^{n \times n}$ is the state transition matrix, $\Gamma_q = h_k \tilde{\theta}_k B_q \in \mathbb{R}^{n \times m}$ is the input matrix, and $S_q = h_k f_q(\tilde{x}_k, \tilde{u}_k^-) \in \mathbb{R}^n$ represents the temporal sensitivity of the state trajectory to the expansion of the pre-event flow duration. Similarly, denoting the complementary post-event time fraction as $1 - \tilde{\theta}_k$, the post-event arc \eqref{eq:Fplus} is linearized as:
\begin{equation} \label{eq:lin_dyn_plus}
    \Delta x_{k+1} = \Phi_r \Delta x_k^+ + \Gamma_r \Delta u_k^+ - S_r \Delta \theta_k,
\end{equation}
where $\Phi_r = (I + h_k (1-\tilde{\theta}_k) A_r)$, $\Gamma_r = h_k (1- \tilde{\theta}_k) B_r$, and $S_r := h_k f_r(\tilde{x}_k^+, \tilde{u}_k^+) \in \mathbb{R}^n$ represents the sensitivity to the corresponding contraction of the post-event flow duration. No division by $\theta_k$ or $(1-\theta_k)$ occurs, as $\tilde{\theta}\rightarrow 1$, \eqref{eq:A+} reduces to $\lambda_k^+=\lambda_{k+1}$ and the only degeneracy is the loss of authority of $u_k^+$ $(\Gamma_r \rightarrow 0)$, regularized by the input penalty in $\mathbf{H}$ \eqref{eq:QP_cost}. Thus we enforce $\theta_k \in [\varepsilon,1-\varepsilon]$, $\varepsilon\ll 1$.

The guard manifold is linearized to enforce geometric consistency at the impact boundary:
\begin{equation} \label{eq:lin_guard}
    \nabla_x g(\tilde{x}_k^-)^\top \Delta x_k^- + \left( h_k \nabla_{t}g(\tilde{x}_k^-) \right) \Delta \theta_k + g(\tilde{x}_k^-, \tilde{\tau}_k) = 0.
\end{equation}
The temporal coefficient uses $\nabla_t g$ rather than $D_t g$ by construction. Substituting \eqref{eq:lin_dyn_minus} in \eqref{eq:lin_guard} makes the total coefficient of $\Delta \theta_k$ equal to $\nabla_x g^{\top}S_q + h_k\nabla_t g = h_k D_tg$. The reduced guard sensitivity is therefore the full time derivative, while the split form \eqref{eq:lin_guard} keeps the flow contribution inside $\Delta x_k^-$ and so avoids double-counting $f_q^-$. The reset map is subsequently linearized as:
\begin{equation} \label{eq:lin_reset}
    \Delta x_k^+ = \nabla_x R(\tilde{x}_k^-) \Delta x_k^- + \left(h_k\nabla_t R(\tilde{x}_k^-)\right)\Delta\theta_k.
\end{equation}

The resulting subproblem minimizes the second-order Gauss-Newton approximation of the Lagrangian subject to the affine constraints:
\begin{equation} \label{eq:QP_cost}
\begin{split}
    &\min_{\Delta z} \quad \frac{1}{2} \Delta z^\top \mathbf{H} \Delta z + \mathbf{q}^\top \Delta z\\
    &\text{subject to:}\; \eqref{eq:lin_dyn_minus}-\eqref{eq:lin_reset}
\end{split}
\end{equation}
where $\mathbf{H}$ is the block-diagonal approximate Hessian and $\mathbf{q}$ is the gradient of the objective \eqref{eq:discret-cost}. 

\subsection{Smoothness and Implicit HMP Sensitivity}

Applying SQP algorithms directly to the implicitly nonsmooth solution maps of standard hybrid integrators violates the Lipschitz continuity requirements for KKT optimality \cite[Ch.~2]{DontchevRockafellar2009}. Standard implementations recover these sensitivities by explicitly computing the saltation matrix $\Xi$ \cite{KongMPC}, whose entries scales as $1/D_t g$ and so degrade as transversality (Eq.~\eqref{eq:transversality}) weakens.

The proposed formulation avoids explicit saltation evaluation by transcribing the hybrid dynamics into a set of smooth algebraic constraints, yielding a $C^1$ NLP representation. This shifts the nonsmooth dependence on event timing into explicit equality constraints, so that standard constraint qualifications hold and SQP iterations remain well-defined \cite{Henrion1992,NocedalWright2006}.

The key mechanism is that event-time optimality is enforced through the dual variables of the KKT system. Let
\begin{equation}
    \mathbf{q}_{\theta_k} = h_k \bigl( (L_{C,q} - L_{C,r}) + \nabla_t L_D \bigr)
\end{equation}
denote the gradient of the objective \eqref{eq:discret-cost} with respect to $\theta_k$, evaluated at the nominal trajectory. The stationarity condition of the QP subproblem yields:
\begin{equation} \label{eq:kkt_timing}
    \nabla_{\theta_k} \mathcal{L}_{QP} = \mathbf{q}_{\theta_k} + \left( h_k \nabla_t g \right)^\top \kappa_k + \left( h_k \nabla_t R \right)^\top \lambda_k^+ = 0.
\end{equation}

The solver enforces \eqref{eq:kkt_timing} by adjusting $\kappa_k$, and the guard and reset duals therefore encode the balance conditions that govern hybrid optimality. Dividing \eqref{eq:kkt_timing} by $h_k$ recovers \eqref{eq:thm_stationarity_proof}, so by Theorem 1 the KKT system is consistent with the HMP without evaluating $\Xi$. No division by $D_t g$ is required to assemble the subproblem \eqref{eq:lin_dyn_minus}-\eqref{eq:lin_reset}. Under Assumption 1 both formulations are well posed and $\kappa_k$ is bounded by Corollary 1.

\subsection{Algorithm and Complexity}
Algorithm \ref{alg:hmp_mpc} summarizes the proposed real-time execution loop. Given the current state $x(t_0)$, the algorithm uses a shifted warm-start of the primal-dual variables from the previous iteration. 

\begin{algorithm}
\caption{Time-Aware Hybrid MPC}\label{alg:hmp_mpc}
\begin{algorithmic}[1]
\State \textbf{Input:} Initial state $x(t_0)$, nominal $\tilde{z}$, duals $(\tilde{\lambda}, \tilde{\kappa})$
\State \textbf{Initialize:} Shift $\tilde{z}$ forward, re-detect $\mathcal{K}$, warm-start.
\For{$k = 0, \dots, N-1$} 
    \State Evaluate Jacobians $A_q, B_q$ and sensitivities $S_q, S_r$.
    \State Construct affine dynamics constraints \eqref{eq:lin_dyn_minus}, \eqref{eq:lin_dyn_plus}.
    \If{transition $\sigma: q \to r$ occurs in interval $k$}
        \State Evaluate $\nabla_x g, \nabla_t g, \nabla_x R, \nabla_t R$ at boundary. 
        \State Construct affine jump constraints \eqref{eq:lin_guard}, \eqref{eq:lin_reset}.
    \EndIf
\EndFor
\State Assemble sparse QP \eqref{eq:QP_cost} evaluating $\mathbf{H}$ and $\mathbf{q}$.
\State Solve QP for primal step $\Delta z$ and updated duals $(\lambda, \kappa)$.
\State \textbf{Update:} $\tilde{z} \gets \tilde{z} + \Delta z$
\State \textbf{Output:} Apply control $u_0^*$ to the system.
\end{algorithmic}
\end{algorithm}

\vspace{0.02in}
\noindent The proposed formulation preserves the $\mathcal{O}(N)$ computational complexity of standard fixed-grid MPC \cite{NocedalWright2006}. Traditional time-aware formulations, such as MPCCs \cite{Baumrucker2009} or global time-scaling methods \cite{KaziWangBiegler2025}, introduce temporal variables that couple across the entire prediction horizon, creating dense borders in the KKT matrix that destroy sparsity. By defining $\theta_k$ as a local, intra-step variable, temporal adjustments are isolated to the $k$-th interval. The resulting KKT matrix retains its block-tridiagonal structure, permitting the use of efficient sparse linear solvers like {\tt OSQP} \cite{osqp}.

\vspace{0.02in}
\noindent Line 2 of Algorithm~\ref{alg:hmp_mpc} re-detects $\mathcal{K}$ from Eq.~\eqref{eq:K} by rolling out the shifted nominal inputs and flagging the intervals in which $g$ changes sign. In each such interval, linear interpolation of $g$ between the two nodes initializes the fraction:
\begin{equation}
    \tilde{\theta} = \frac{-g(\tilde{x}_k,t_k)}{g(\tilde{x}_{k+1},t_{k+1})-g(\tilde{x}_k,t_k)}.
\end{equation}
Because $\theta_k$ is confined to its own interval, an event that drifts past a grid node drives $\tilde{\theta}_k$ to a bound. The transition block is then moved to the adjacent interval, with $(\theta_k,\kappa_k)$ reused as its warm start. The decision vector is permanently sized for $|\mathcal{K}|_{\max}$ blocks, so the sparsity pattern never changes and the solver is never re-set up.

At most one transition per interval is admitted. If the rollout detects two crossings in one interval, that interval reverts to a fixed-grid saltation step \cite{KongMPC}. Thus the QP stays feasible and its structure is unchanged, and only intra-step timing optimality is lost, on that interval alone. Handling several genuine intra-step transitions, as in multi-contact locomotion, requires multiple $\theta_k$ per interval and remains future work.

As with RTI schemes generally, solving one QP per cycle give up the guarantees of a converged SQP, and we provide no closed-loop stability certificate. Nominal stability conditions for RTI-MPC are available under contraction and shift-initialization assumptions \cite{DiehlStability2005}. Extending them to the lifted hybrid setting, where $\mathcal{K}$ may change between cycles, remains open.

%% file: 5-NumericalResults.tex
\begin{figure*}[t!]
    \centering
    \begin{subfigure}[b]{0.67\linewidth}
        \centering
        \includegraphics[width=\linewidth]{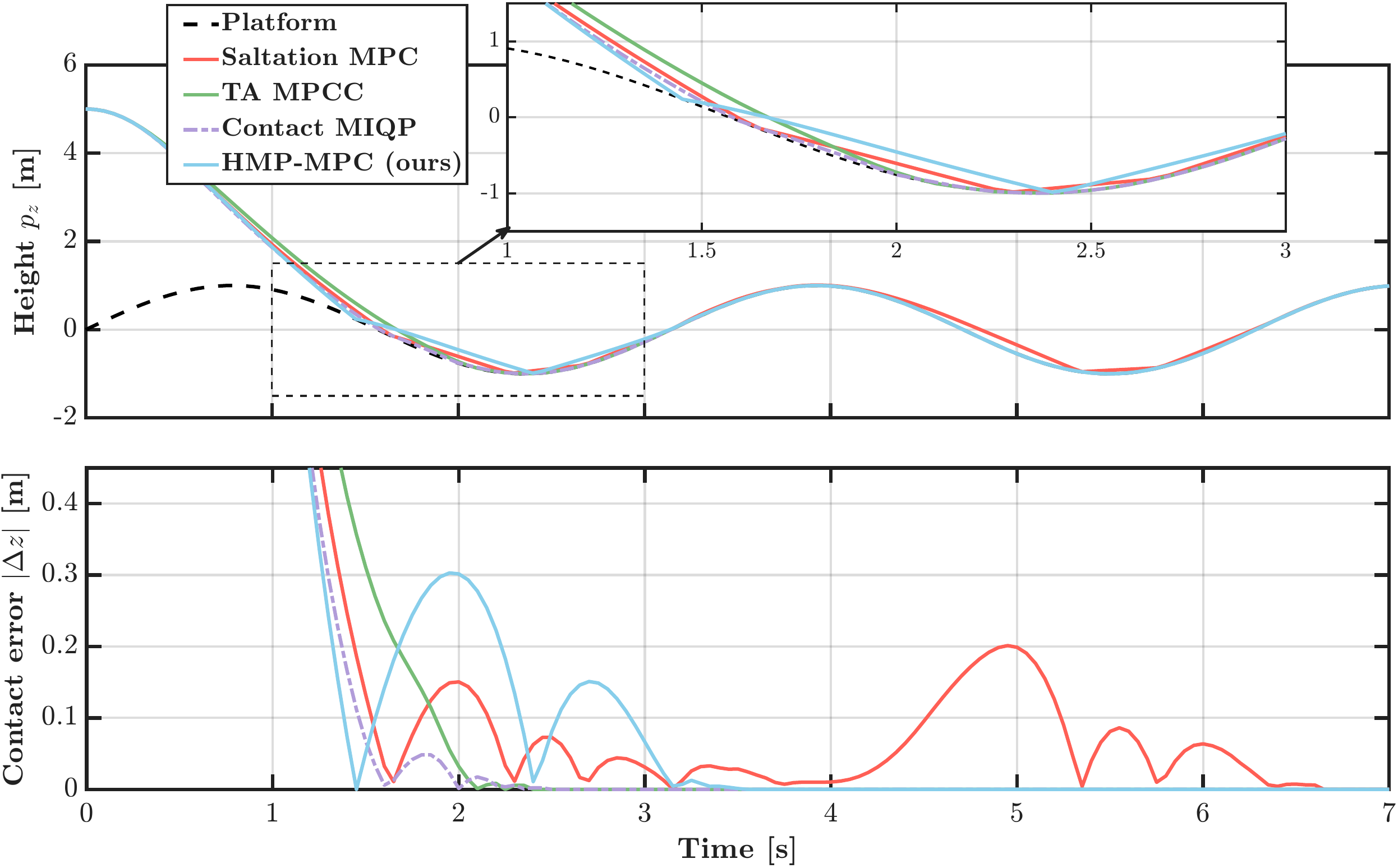}
        \caption{Tracking performance of the four studied methods.}
        \label{fig:Performance}
    \end{subfigure}%
    \hfill
    \begin{minipage}[b]{0.28\linewidth}
        \centering
        \begin{subfigure}{\linewidth}
            \centering
            \includegraphics[width=\linewidth]{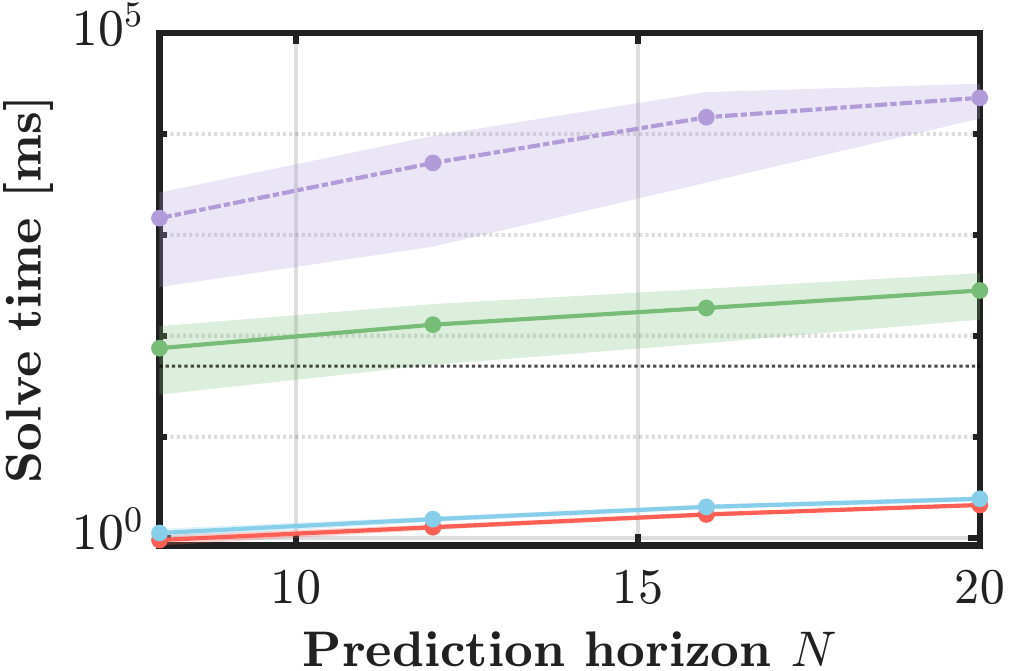}
            \caption{Per-cycle solve time against $N$.}
            \label{fig:Computation}
        \end{subfigure}
        \vspace{1.0em}
        \begin{subfigure}{\linewidth}
            \centering
            \includegraphics[width=\linewidth]{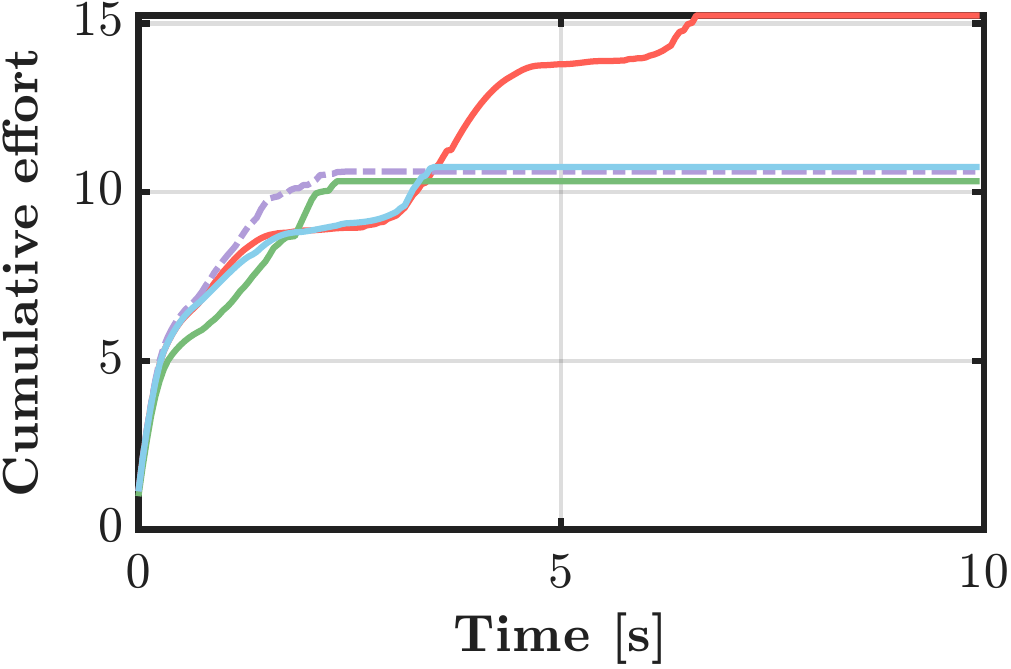}
            \caption{Control Effort of 4 methods.}
            \label{fig:Cost}
        \end{subfigure}%
    \end{minipage}
    \caption{Closed-loop landing on a moving platform comparison on the four studied methods.}
    \label{fig:Results}
\end{figure*}

\begin{table*}
\centering
\caption{Closed-loop performance and per-cycle cost ($N=20$, $h=0.05$~s). Solve times are measured on identical sampled linearization points, so solvers visiting different closed-loop states remain comparable. The MIQP is capped at 30~s per solve.}
\label{tab:combined_metrics}
\begin{tabular}{@{}l c ccc c ccc@{}}
\toprule
\textbf{Method / solver} & & \textbf{Attach (s)} & \textbf{Impacts} & \textbf{Cum. effort} & & \textbf{Solve time (ms)} & \textbf{Worst (ms)} & \textbf{Rel. cost} \\
\midrule
Saltation MPC (OSQP) & \multirow{5}{*}{\rotatebox[origin=c]{90}{\textbf{Performance}}} & 6.35 & 6 & 15.2 & \multirow{5}{*}{\rotatebox[origin=c]{90}{\textbf{Computation}}} & 2.11 $\pm$ 0.07 & 2.26 & 0.9$\times$ \\
\textbf{HMP-MPC (OSQP)} &  & 3.20 & 4  & 10.7 &  & 2.43 $\pm$ 0.14 & 2.72 & 1.0$\times$ \\
HMP-MPC (IPOPT) &  & 3.25 & 4  & 11.4 &  & 26.6 $\pm$ 9.7 & 35.3 & 11$\times$ \\
Time-aware MPCC (IPOPT) &  & 2.10 & 2  & 10.3 &  & 281 $\pm$ 136 & 554 & 116$\times$ \\
Contact MIQP (Bonmin) &  & 2.05 & 5  & 10.6 &  & 22\,800 $\pm$ 8\,600 & 30\,400 & 9383$\times$ \\
\bottomrule
\end{tabular}
\end{table*}

\section{Simulation Results}

\vspace{0.02in}
\noindent \textbf{Simulation Setup:}
Using hybrid system benchmarks \cite{KongMPC, KaziWangBiegler2025, MIQPSanfelice}, we simulate a 3D controlled bouncing mass interacting with a moving platform. The system state is defined as $x = [p, v]^{\top} \in \mathbb{R}^6$, representing the 3D position and velocity. The continuous-time dynamics are given by the vector field $f(x, u) = [v, u - g_v]^T$, where $u \in \mathbb{R}^3$ is the control input ($u_z\in[0,15]$) and $g_v = [0, 0, g]^T$ is the gravity vector.

An impact occurs when the system intersects the guard function $g(x, t) = \Delta z(t) =  p_z(t) - z_p(t) = 0$. Following standard rigid-body impact models \cite{leine2003}, the normal velocity reflects relative to the platform with a restitution coefficient $e=0.75$, and the tangential friction $\mu_t=0.2$. The jump map $R(x, t)$ resets the state to:
\begin{equation}
    R:
    \begin{cases}
      p^+&=p^-\\
      v^+ &= [\mu_t v_x^-,\  \mu_t v_y^-,\ v_p(t) -e(v_z^- - v_p(t))]^{\top}.
    \end{cases}
\end{equation}
The optimization minimizes a quadratic continuous flow cost $L_{C,q}=L_{C,r}=\|x - x_{ref}\|_Q^2 + \|u\|_R^2$ over the prediction horizon $N$. Because the impact does not incur an instantaneous energy penalty, the discrete jump cost is set to zero ($L_D \equiv 0$). This instantiates \eqref{eq:discret-cost} with:
\begin{equation} \label{eq:cost_func}
    \mathcal{J} = \sum_{k=0}^{N-1} L_{C,q}(x_k,u_k) + \|x_N - x_{ref}\|_{Q_N}^2
\end{equation}
where $Q$, $R$, and $Q_N$ are positive definite weight matrices, and any interval containing a jump evaluates the split continuous cost $L_{C,\sigma_k}$ accordingly. The mass is declared \emph{attached} once it stays within $1$ cm of the surface for $0.3$ s.

The platform follows $z_p(t)=A\sin(\omega t)$, $A=1$ and $\omega=2$ rad/s, and all controllers use $dt=0.05$ s and $N=20$. We compare four formulations, summarized in Table \ref{tab:combined_metrics}. 

\emph{Saltation MPC} \cite{KongMPC} is a fixed-step sparse QP (using {\tt OSQP} \cite{osqp}) that propagates sensitivities via the saltation matrix but commits the reset to a grid node. \emph{Time-aware (TA) MPCC} \cite{Baumrucker2009} is a complementarity-constrained NLP solved by {\tt IPOPT} \cite{IPOPT} through {\tt CasADi} \cite{casadi}. \emph{Contact MIQP} \cite{MIQPSanfelice} is a big-M mixed-logical-dynamical transcription with one binary per interval selecting whether an impact occurs, solved by branch and bound ({\tt Bonmin}~\cite{BONAMI2008}). 

\emph{HMP-MPC (ours)} implemented as a sparse QP using {\tt OSQP}. To separate the effect of the transcription from that of the solver, we additionally solve \emph{the same} lifted formulation as a general NLP with {\tt IPOPT}.

\vspace{0.02in}
\noindent \textbf{Tracking Performance Results:}
All four controllers reach the platform, but they differ in how directly they get there (Fig. \ref{fig:Results}). The fixed-grid saltation MPC cannot place the reset inside a step, so it consistently mistimes the impact: its one-step prediction error across an impact-containing interval averages $7.9$ cm, against $0.7$ cm for the lifted transcription. Because that error is larger than the attachment band itself, the baseline overshoots, rebounds, and requires $6$ impacts and $6.35$ s to attach. HMP-MPC anticipates the impact instant, approaches with near-zero relative velocity, and attaches in $3.20$ s using $4$ impacts. The two offline-quality formulations attach soonest ($2.10$ s for TA MPCC, $2.05$ s for the MIQP), as expected of methods that solve the nonlinear or combinatorial problem to (near) optimality. The relevant comparison is what that costs, discussed next.

\begin{figure}[h!]
    \centering
    \includegraphics[width=\linewidth]{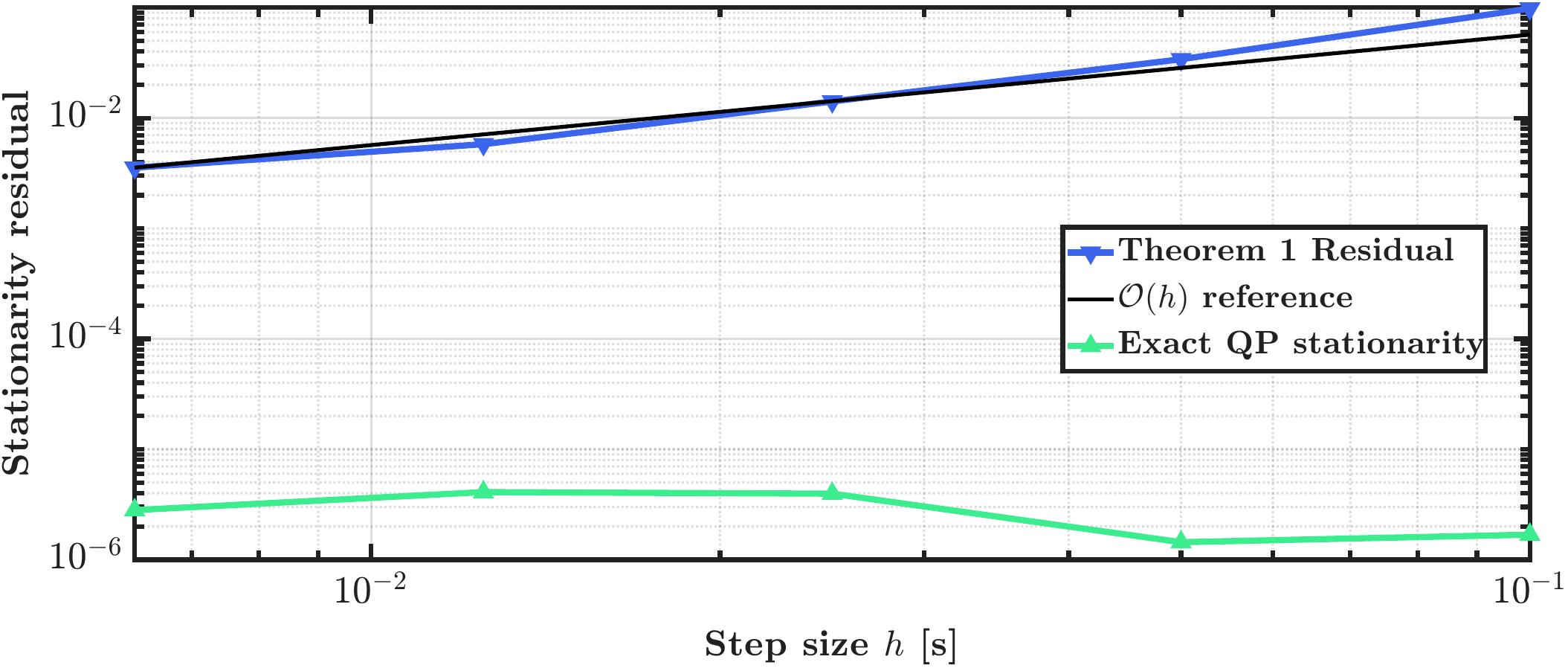}
    \caption{Stationarity residual of Eq.~\eqref{eq:thm_stationarity_proof} against step size $h$.
    The QP stationarity residual stays at solver tolerance.}
    \label{fig:stationarity}
\end{figure}

\vspace{0.02in}
\noindent\textbf{Discretization Consistency:} Theorem~1 bounds the gap between the discrete KKT system and the continuous-time HMP by $\mathcal{O}(h_k)$, but at the fixed step used here ($h=0.05$~s) that remainder does not vanish. Fig.~\ref{fig:stationarity} evaluates the residual of~\eqref{eq:thm_stationarity_proof} at the converged solution over
$h\in[10^{-2},10^{-1}]$. The residual decays with an empirical slope of $1.21$, consistent with the first-order bound, while the QP stationarity residual remains at solver tolerance throughout, confirming that the decay reflects discretization error rather than solver accuracy. At $h=0.05$~s the residual is $\approx 3\times 10^{-2}$, small relative to the cost scale, indicating that the attach-time gap against the offline solvers in Table~\ref{tab:combined_metrics} stems from the single-QP RTI approximation rather than from truncation error. This gives a direct criterion for selecting $h$, as it may be reduced until the residual falls below the accuracy demanded by the task, subject to the linear growth of solve time in $N$ (Fig. \ref{fig:Computation}).

\vspace{0.02in}
\noindent \textbf{Energy and Computational Analysis:}
Because the formulation enforces event-time stationarity, the controller aligns impact timing with the optimality conditions. By optimizing the intra-step timing, HMP-MPC eliminates the unmodeled rebounds that dominate the fixed-grid trajectory and drives the active control effort to zero once attached. Against the only other real-time-capable formulation, this yields a $30\%$ reduction in cumulative effort ($10.7$ vs $15.2$) and a third fewer impacts, for a $15\%$ increase in solve time (Fig~\ref{fig:Cost}).

The cost of each formulation is separated in Table \ref{tab:combined_metrics}. Comparing the two {\tt IPOPT} rows isolates the transcription, the lifted formulation is $10.6\times$ cheaper than the complementarity expansion at the same solver. Comparing the two HMP-MPC rows isolates the solver, and a further $10.9\times$, obtained because the lifted transcription remains a convex QP with linear constraints and can therefore be handed to a first-order sparse QP solver, which the complementarity constraints prevent. Together these give the $116\times$ end-to-end speedup over TA MPCC. The MIQP returns the best mode sequence found and lands well, but at $22.8$ s per cycle, some $9400\times$ the proposed method and three orders of magnitude beyond the sampling period. Fig. \ref{fig:Computation} shows the formulations computational cost scaling in $N$.

\vspace{0.05in}
\noindent \textbf{Sensitivity to Contact Restitution:}
We sweep the plant's true restitution $e_{\text{true}}$ over $[0,0.95]$ with $40$ randomized initial conditions per value, and run the sweep twice. Once giving each controller the true value (\emph{exact}), once holding it at the nominal $e_{\text{MPC}}=0.75$ (\emph{mismatched}). The first isolates the difficulty of the task, the second adds the cost of an inaccurate contact model. No parameters are retuned.

Below $e_{\text{true}}=0.60$ every method captures in all trials, since an impact that dissipates more energy than modeled reaches the attachment condition sooner. Thus values under $e_{\text{true}}=0.60$ are not reported in Table.\ref{tab:robustness}.

The \emph{exact} columns show that what fails on a bouncy surface is timing, not modeling. Both controllers know the contact physics exactly, yet the fixed-grid baseline misses roughly a third of its landings at $e_{\text{true}}\ge0.85$ while HMP-MPC captures $98$--$100\%$. A larger $e$ amplifies any error in the predicted impact instant, because the post-impact trajectory diverges faster from the mistimed one. That error is the $\mathcal{O}(h_k)$ event localization of the fixed grid, and removing it is exactly what lifting $\theta_k$ does.

The $\Delta$ columns show that the two formulations also absorb model error differently. HMP-MPC stays within the confidence interval at every $e_{\text{true}}$, whereas the baseline loses $26$ points at $0.95$. Minimizing over $\theta_k$ brings the mass to contact at 
near-zero relative velocity, and since $v^+=-e\,v^-$, the model error enters only in proportion to $v^-$. Arriving slowly attenuates the restitution error rather than leaving the controller to reject it afterwards.

\begin{table}
\centering
\caption{Capture rate (\%) over $40$ randomized initial conditions per cell.
\emph{Exact} gives each controller the plant's true restitution
$e_{\text{true}}$. \emph{Mismatched} holds its model at the nominal
$e_{\text{MPC}}=0.75$, and $\Delta$ is the cost of that mismatch.}
\label{tab:robustness}
\begin{tabular}{@{}lccccccc@{}}
\toprule
 & \multicolumn{3}{c}{\textbf{Saltation MPC}} & & \multicolumn{3}{c}{\textbf{HMP-MPC}} \\
\cmidrule(lr){2-4}\cmidrule(lr){6-8}
$e_{\text{true}}$ & exact & mism. & $\Delta$ & & exact & mism. & $\Delta$ \\ \midrule
$0.60$ & 100 & 100 & $0$   & & \textbf{100} & \textbf{100} & $0$  \\
$0.75$ &  95 &  95 & $0$   & & \textbf{100} & \textbf{100} & $0$  \\
$0.85$ &  65 &  62 & $-3$  & & \textbf{100} & \textbf{98}  & $-2$ \\
$0.95$ &  68 &  42 & $-26$ & & \textbf{98}  & \textbf{95}  & $-3$ \\
\bottomrule
\end{tabular}
\end{table}

%% file: 6-Conclusion.tex
\section{Conclusion and Future Work}

This paper presented time-aware Hybrid Model Predictive Control (HMP-MPC). Rather than evaluating the saltation matrices numerically, the formulation embeds HMP time-stationarity conditions directly into the NLP and captures first-order hybrid sensitivities through the KKT system. Validated on a dynamic landing task, HMP-MPC optimizes the rebound timing, reducing post-impact control effort and attach times relative to fixed-grid saltation MPC without the computational burden of MPCC. 

The validation is limited to a single impact per horizon on a low-dimensional benchmark. Extending HMP-MPC to multi-contact/higher-dimensional regimes with several intra-step transitions, is the subject of ongoing work. We will also extend the controller design to handle parametric uncertainties such as unknown restitution coefficients, and validate real-time feasibility on physical hardware.